\documentclass{article}
\usepackage{spconf,amsmath,graphicx}

\let\OLDthebibliography\thebibliography
\renewcommand\thebibliography[1]{
  \OLDthebibliography{#1}
  \setlength{\parskip}{0pt}
  \setlength{\itemsep}{0pt plus 0.4ex}
}

\title{Skin Lesion Analysis toward Melanoma Detection:
A Challenge at the 2017 International Symposium on Biomedical Imaging (ISBI), hosted by the International Skin Imaging Collaboration (ISIC)}
\begin{document}
 \name{ \em  Noel C. F. Codella$^{1\dagger}$, David Gutman$^{2\dagger}$,  M. Emre Celebi$^{3}$, Brian Helba$^{4}$, \\ \em Michael A. Marchetti$^{5}$, Stephen W. Dusza$^{5}$,  Aadi Kalloo$^{5}$, Konstantinos Liopyris$^{5}$, \\ \em  Nabin Mishra$^{6}$, Harald Kittler$^{7}$, Allan Halpern$^{5\ddagger}$  
 \thanks{$^{\dagger}$ The first two authors contributed equally to this work.} \\
\thanks{$^{\ddagger}$ Corresponding author.}
\thanks{Accepted for publication at ISBI 2018. \copyright 2018 IEEE}
}

 \address{$^{1}$ IBM T. J. Watson Research Center, Yorktown Heights, NY, USA\\ $^{2}$ Emory University, Atlanta, GA, USA \\  $^{3}$ University of Central Arkansas, Conway, AR, USA \\ $^{4}$ Kitware, Clifton Park, NY, USA \\ $^{5}$ Memorial Sloan-Kettering Cancer Center, New York, NY, USA \\ $^{6}$ Missouri University of Science and Technology, Rolla, MO USA \\ $^{7}$ Medical University of Vienna, Vienna, Austria }

%
\maketitle
%
\begin{abstract}
This article describes the design, implementation, and results of the latest installment of the dermoscopic image analysis benchmark challenge. The goal is to support research and development of algorithms for automated diagnosis of melanoma, the most lethal skin cancer. The challenge was divided into 3 tasks: lesion segmentation, feature detection, and disease classification. Participation involved 593 registrations, 81 pre-submissions, 46 finalized submissions (including a 4-page manuscript), and approximately 50 attendees, making this the largest standardized and comparative study in this field to date. While the official challenge duration and ranking of participants has concluded, the dataset snapshots remain available for further research and development.

\end{abstract}

\section{Introduction}
\label{sec:intro}


The most prevalent form of cancer in the United States is skin cancer, with 5 million cases occurring annually \cite{rate,cancerfacts,10k}. Melanoma, the most dangerous type, leads to over 9,000 deaths a year ~\cite{cancerfacts,10k}. Even though most melanomas are first discovered by patients \cite{patients}, the diagnostic accuracy of unaided expert visual inspection is only about 60\% \cite{clinicalaccuracy1}. 

Dermoscopy is a recent technique of visual inspection that both magnifies the skin and eliminates surface reflection.  Research has shown that with proper training, diagnostic accuracy with dermoscopy is 75\%-84\% \cite{clinicalaccuracy1,clinicalaccuracy2,MM01}. In an attempt to improve the scalability of dermoscopic expertise, procedural algorithms, such as ``3-point checklist,'' ``ABCD rule,'' ``Menzies method,'' and ``7-point checklist,'' were developed  ~\cite{clinicalaccuracy2,netmeeting}. However, many clinicians forgo these methods in favor of relying on personal experience, as well as the ``ugly duckling'' sign (outliers on patient) ~\cite{clinicalprocess}. 

Recent reports have called attention to a growing shortage of dermatologists per capita ~\cite{shortage}.  This has increased interest in techniques for automated assessment of dermoscopic images
\cite{nabin,review,codellajrd,ph2}. 
However, most studies have used isolated silos of data for analysis that are not available to the broader research community. While an earlier effort to create a public archive of images was made \cite{ph2,ph2dataset}, the dataset was too small (200 images) to fully represent scope of the task. 

The International Skin Imaging Collaboration (ISIC) has begun to aggregate a large-scale publicly accessible dataset of dermoscopy images.  Currently, the dataset houses more than 20,000 images from leading clinical centers internationally, acquired from a variety of devices used at each center. The ISIC dataset was the foundation for the first public benchmark challenge on dermoscopic image analysis in 2016 ~\cite{challenge,jaadarticle}. The goal of the challenge was to provide a fixed dataset snapshot to support development of automated melanoma diagnosis algorithms across 3 tasks of lesion analysis:  segmentation, dermoscopic feature detection, and classification.  


In 2017, ISIC hosted the second instance of this challenge, featuring an expanded dataset. In the following sections, the datasets, tasks, metrics, participation, and the results of this challenge are described.

\section{DATASET DESCRIPTIONS \& TASKS}
\label{sec:datasets}

The 2017 challenge consisted of 3 tasks: lesion segmentation, dermoscopic feature detection, and disease classification. For each, data consisted of images and corresponding ground truth annotations, split into training (n=2000), validation (n=150), and holdout test (n=600) datasets.  Predictions could be submitted on validation and test datasets. The validation submissions provided instantaneous feedback in the form of performance evaluations, as well as ranking in comparison to other participants. Test submissions only provided feedback after the submission deadline. The training, validation, and test datasets continue to be available for download from the following address: http://challenge2017.isic-archive.com/

\begin{figure}[t]
  \centerline{\includegraphics[width=6.5cm]{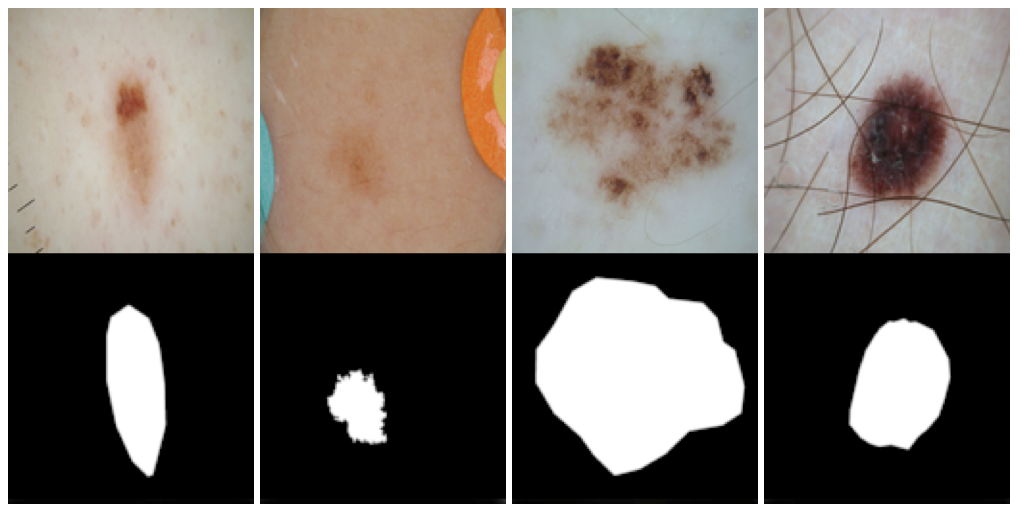}}
  \caption{Images from ``Part 1: Lesion Segmentation.'' {\em Top:} Original images. {\em Bottom:} Segmentation masks.   }
\label{fig:part1}
\end{figure}

\begin{figure}[t]
  \centerline{\includegraphics[width=6.5cm]{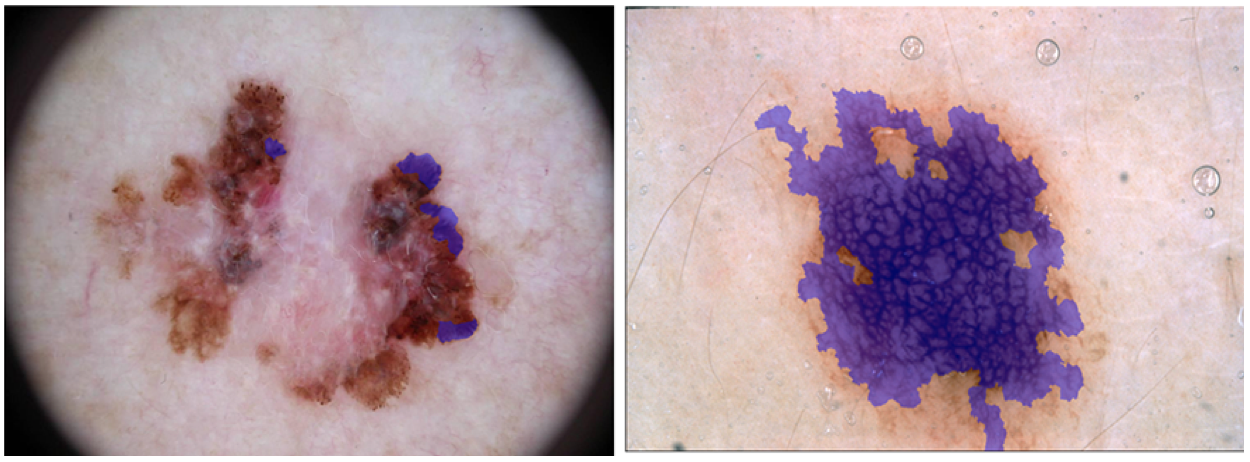}}
  \caption{Images from ``Part 2: Dermoscopic Feature Classification''. Ground truth labels highlighted in purple. {\em Left:} Streaks. {\em Right:} Pigment Network.    }
\label{fig:part2}
\end{figure}

\begin{figure}[t]
  \centerline{\includegraphics[width=8cm]{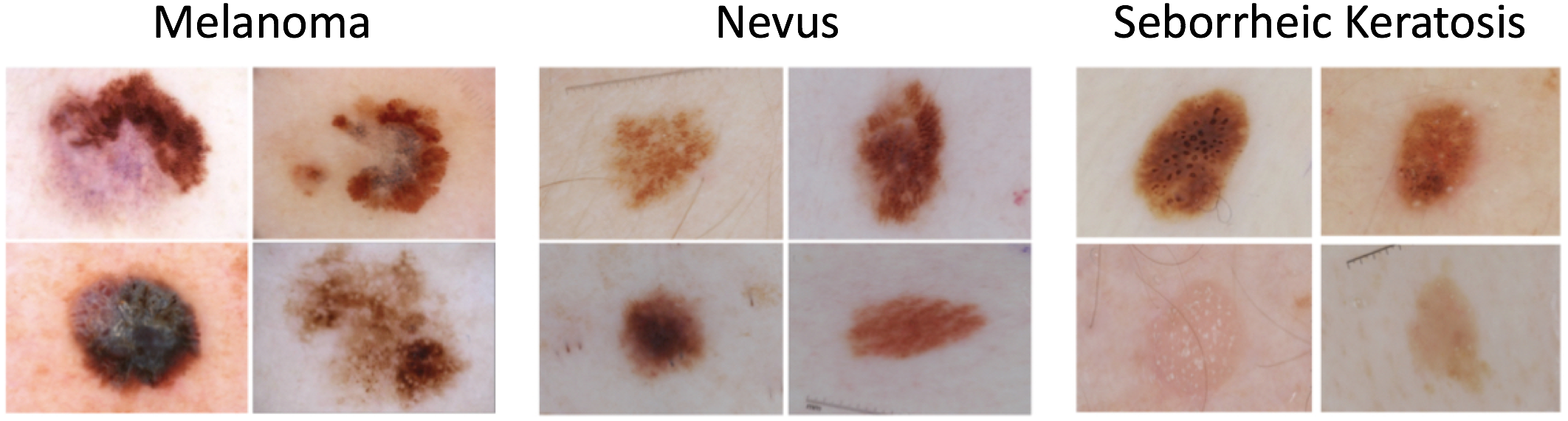}}
  \caption{Example images from ``Part 3: Disease Classification.'' Ground truth labels written above. }
\label{fig:part3}
\end{figure}

{\bf Part 1: Lesion Segmentation Task: } Participants were asked to submit automated predictions of lesion segmentations from dermoscopic images in the form of binary masks. Lesion segmentation training data included the original image, paired with the expert manual tracing of the lesion boundaries also in the form of a binary mask, where pixel values of 255 were considered inside the area of the lesion, and pixel values of 0 were outside (Fig. ~\ref{fig:part1}). 

{\bf Part 2: Dermoscopic Feature Classification Task: } Participants were asked to automatically detect the following four clinically defined dermoscopic features: ``network,'' ``negative network,'' ``streaks,'' and ``milia-like cysts,'' ~\cite{clinicalterms1,clinicalterms2}. Pattern detection involved both localization and classification (Fig. \ref{fig:part2}). To reduce the variability and dimensionality of spatial feature annotations, the lesion images were subdivided into superpixels using the SLIC algorithm \cite{slic}. Lesion dermoscopic feature data included the original lesion image and a corresponding set of superpixel masks, paired with superpixel-wise expert annotations for the presence or absence of the dermoscopic features. Validation and test sets included images and superpixels without annotation.

{\bf Part 3: Disease Classification Task: } Participants were asked to classify images as belonging to one of 3 categories (Fig. ~\ref{fig:part3}), including ``melanoma'' (374 training, 30 validation, 117 test), ``seborrheic keratosis'' (254, 42, and 90), and ``benign nevi'' (1372, 78, 393), with classification scores normalized between 0.0 to 1.0 for each category (and 0.5 as binary decision threshold). Lesion classification data included the original image paired with the gold standard diagnosis, as well as approximate age (5 year intervals) and gender when available. 

\section{EVALUATION METRICS}



Details of evaluation metrics have been previously described ~\cite{challenge,jaadarticle}. For classification decisions, any confidence above 0.5 was considered positive for a category. For segmentation tasks, pixel values above 128 were considered positive, and pixel values below were considered negative.

For evaluation of classification decisions, the area under curve (AUC) measurement from the receiver operating characteristic (ROC) curve was computed ~\cite{challenge}.



Additionally, for classification of melanoma, specificity was measured on the operating curve where sensitivity was equal to 82\%, 89\%, and 95\%, corresponding to dermatologist classification and management performance levels, and theoretically desired sensitivity levels, respectively ~\cite{jaadarticle}. 



Segmentation submissions were compared using the Jaccard Index, Dice coefficient, and pixel-wise accuracy \cite{challenge}.  Participant ranking used Jaccard. 


%
%
%



\section{RESULTS}
\label{sec:results}

\begin{figure}[t]
  \centerline{\includegraphics[width=8.5cm]{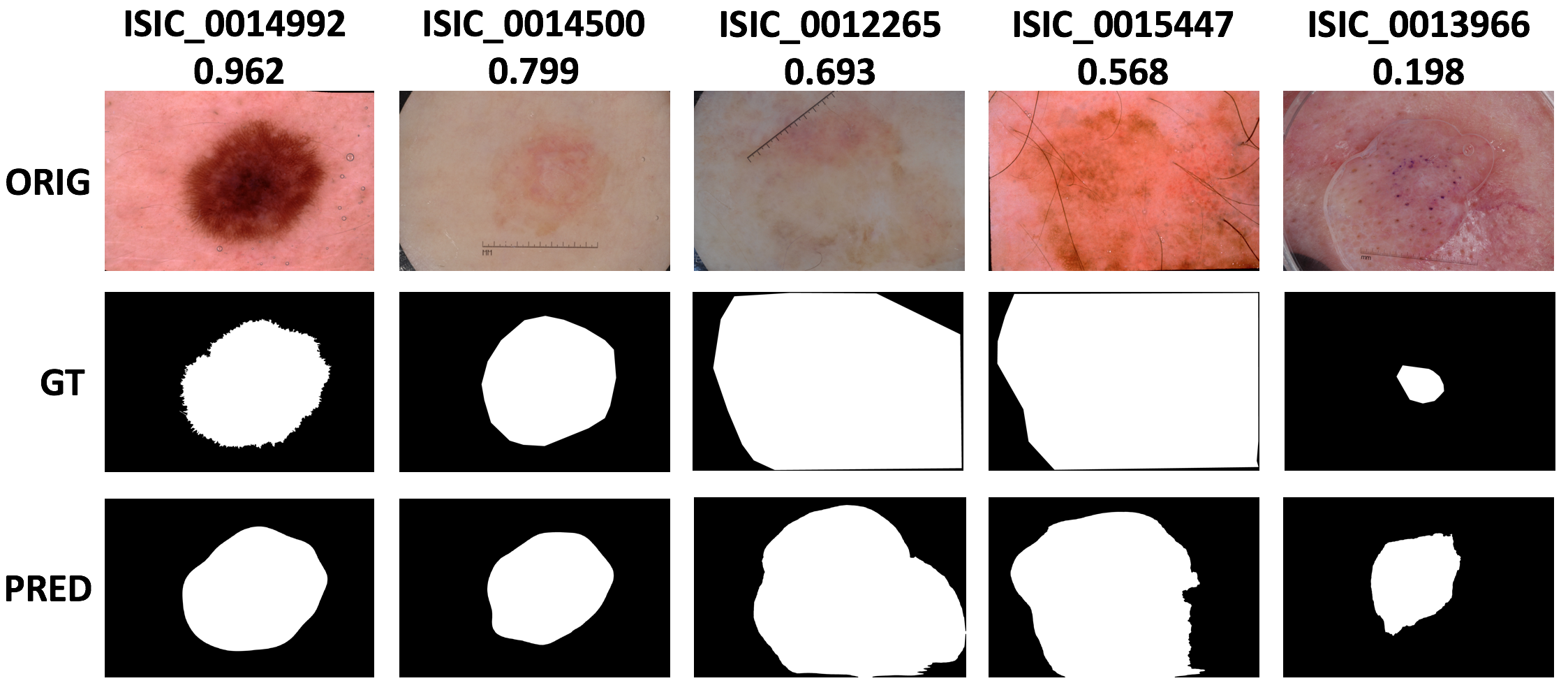}}
  \caption{Part 1 example segmentations from top ranked participant submission. {\em Top Row:} Original images.  {\em Middle Row:} Ground truth segmentations. {\em Bottom Row:} Participant predictions. ISIC identifiers and Jaccard Index values are listed at each column head.   }
\label{fig:part1grid}
\end{figure}

\begin{figure}[t]
  \centerline{\includegraphics[width=8.5cm]{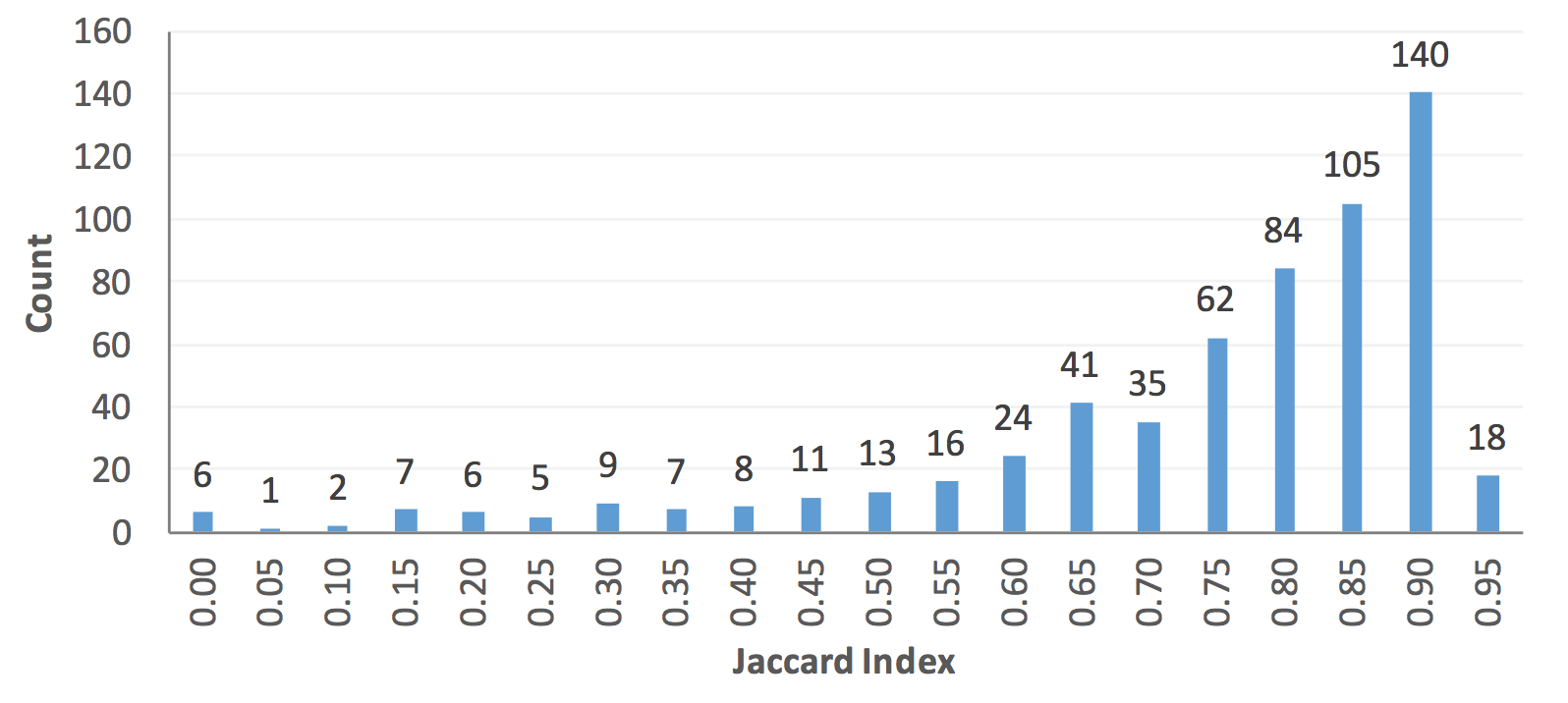}}
  \caption{Histogram of Jaccard Index values for individual images from top segmentation task participant submission. }
\label{fig:part1hist}
\end{figure}

The 2017 challenge saw 593 registrations, 81 pre-submissions, and 46 finalized submissions (including a 4 page arXiv paper with each). The associated workshop at ISBI 2017 saw approximately 50 attendees. To date, this has been the largest standardized and comparative study in this field, accounting for the size of the dataset, the number of algorithms evaluated, and the number of participants. In the following, the results for each challenge part are investigated.

{ \bf Part 1: Lesion Segmentation Task: }21 sets of prediction scores on the final test set were submitted for the segmentation task, and 39 were submitted to the validation set. The top ranked participant achieved an average Jaccard Index of 0.765, accuracy of 93.4\%, and Dice coefficient of 0.849, using a variation of a fully convolutional network ensemble (a deep learning approach) ~\cite{part1rank1}. Example segmentations are shown in Fig. ~\ref{fig:part1grid}, and a histogram of individual image Jaccard Index measurements is shown in Fig. ~\ref{fig:part1hist}. Subjectively assessing the quality of the segmentations, one can observe that segmentations of Jaccard Index 0.8 or above tend to appear visually ``correct.'' This observation is consistent with prior reports that measured an inter-observer agreement of 0.786 on a subset of 100 images from the ISIC 2016 Challenge ~\cite{codellajrd}. When Jaccard falls to 0.7 or below, the ``correctness'' of the segmentation can be debated. 156 out of 600 images (26\%) fell at or below a Jaccard of 0.7. 91 images (15.2\%) fell at or below Jaccard of 0.6. This suggests a failure rate of 15\% to 26\%, which is higher than the pixel-wise failure rate of 6.6\%. 

\begin{table}
\centering
\begin{tabular}{|p{1.cm}|p{1.cm}|p{1.cm}|p{1.cm}|p{1.cm}|p{1.cm}|} 
 \hline
 \bf Method / Rank & \bf AVG & \bf Net- work & \bf Neg. Network & \bf Streaks & \bf Milia-Like Cyst  \\ 
 \hline 
 \bf [22] / 1 & \bf 0.895 & \bf 0.945 & \bf 0.869 & \bf 0.960 & 0.807 \\ 
 \hline
 \bf [23] / 2 & 0.833 & 0.835 & 0.762 & 0.896 & \bf 0.838  \\ 
 \hline
 \bf [23] / 3 & 0.832 & 0.828  & 0.762 & 0.900 & 0.837  \\
 \hline
\end{tabular}
\caption{Part 2: Dermoscopic Feature Classification AUC Measurements. AVG = Average across all categories.  }
\label{table:segresults}
\end{table}

\begin{table*}[!htbp]
\centering
\begin{tabular}{|p{2.5cm}|p{1.cm}|p{1.cm}|p{1.cm}|p{1.cm}|p{1.cm}|p{1.cm}|p{1.cm}|p{1.cm}|p{1.cm}|p{1.cm}|} 
 \hline
 \bf Method & \bf AVG-AUC & \bf M-AUC & \bf SK-AUC  & \bf M-SP82 & \bf M-SP89 & \bf M-SP95 & \bf M-SENS & \bf M-SPEC & \bf SK-SENS & \bf SK-SPEC \\ 
 \hline 
  \bf [24] Top AVG & 0.911 & 0.868 & 0.953 &  0.729 & 0.588 & 0.366 & 0.735 & 0.851 & 0.978 & 0.773\\ 
 \hline
  \bf [25] Top SK & 0.910 & 0.856 & 0.965 & 0.727 & 0.555 & 0.404 & 0.103  & 0.998 & 0.178 & 0.998 \\ 
 \hline
  \bf [26] Top M & 0.908 & 0.874 & 0.943 & 0.747 & 0.590 & 0.395 & 0.547 & 0.950 & 0.356 & 0.990 \\
 \hline
 \bf AVGSC & 0.913 & 0.872 & 0.954 & 0.778 & 0.605 & 0.435 & 0.214 & 0.988 & 0.600 & 0.975 \\
 \hline
 \bf L-SVM & 0.926 & 0.892 & 0.960 & 0.834 & 0.692 & 0.571& 0.718 & 0.901 & 0.878 & 0.931 \\
 \hline
 \bf NL-SVM & 0.904 & 0.853 & 0.955 & 0.801 & 0.449 & 0.168 & 0.675 & 0.909 & 0.889 & 0.928 \\
 \hline
\end{tabular}
\caption{Part 3: Disease classification evaluation metrics for top 3 participants, followed by average score, linear SVM, and non-linear SVM 3-fold cross-validation fusion methods. {\em Key:} M = Melanoma. SK = Seborrheic Keratosis. AVG = Average between two classes. AUC = Area Under Curve. SP82/89/95 = Specificity mearured at 82/89/95\% sensitivity. L-SVM = Linear SVM. NL-SVM = Non-Linear SVM.  AVGSC = Average Score. }
\label{table:classresults}
\end{table*}

{\bf Part 2: Dermoscopic Feature Classification Task:} 
For the second year in a row, dermoscopic feature classification has received far less participation than other tasks. Only 3 submissions ~\cite{part2rank1,part2rank2} on the test set were received from 2 parties. Whether this is due to the technical framing of the task (how well it maps to existing frameworks), or the perceived importance of the task, is a matter of current investigation. 

Regardless, performance levels of those submissions that were received demonstrated that localization of dermoscopic features is a tractable task for computer vision approaches. Top performance levels are shown in  Table ~\ref{table:segresults}. AUC was above 0.75 ubiquitously, with an average close to 0.9. 

\begin{figure}[t]
  \centerline{\includegraphics[width=9cm]{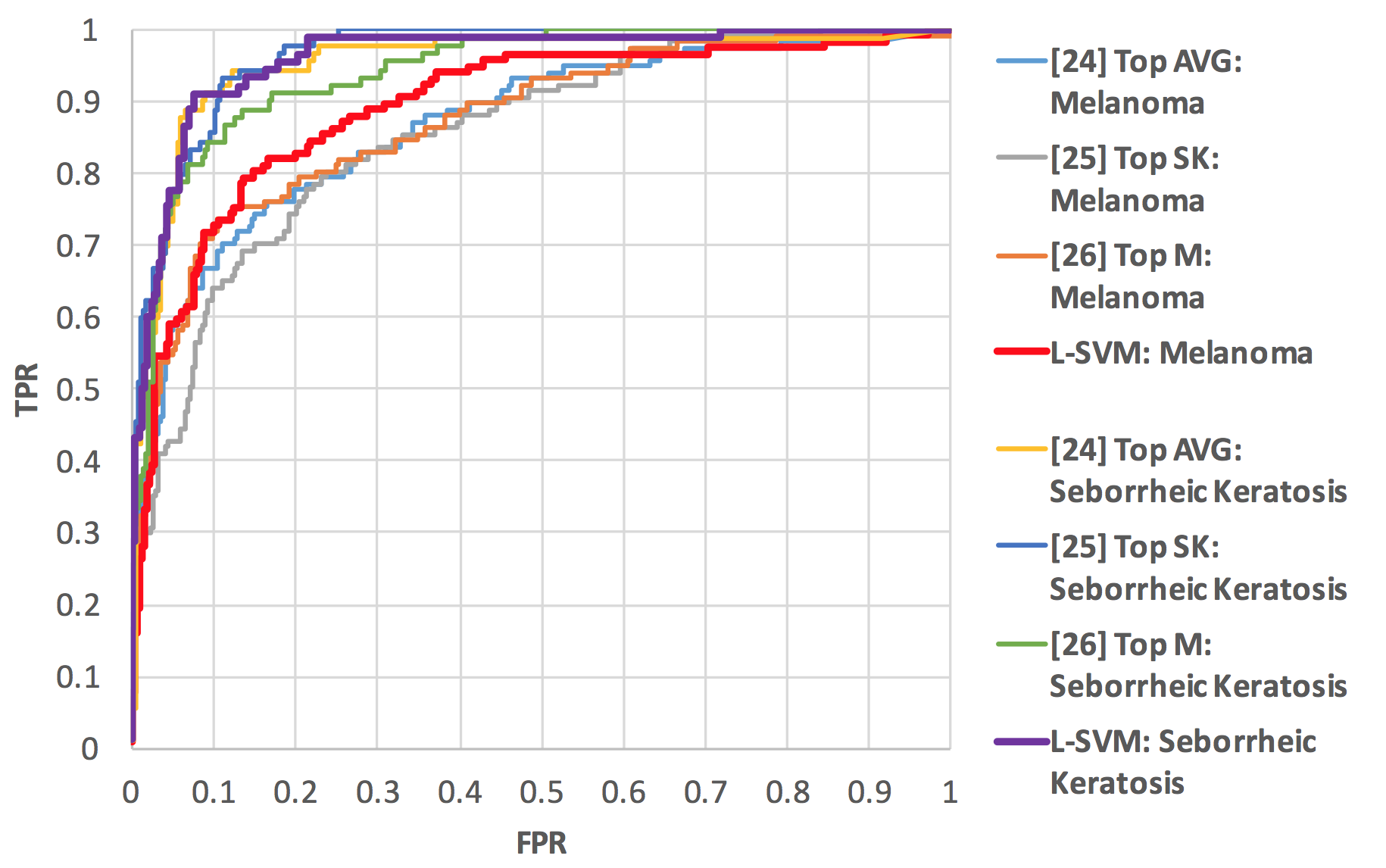}}
  \caption{ROC curves for top 3 submissions to ``Part 3: Disease Classification'', as well as linear SVM fusion. }
\label{fig:part3roc}
\end{figure}

{\bf Part 3: Disease Classification Task: }The disease classification task received 23 final test set submissions, and 39 validation set submissions. Performance characteristics of the average (AVG) classification winner ~\cite{casio}, seborrheic keratosis (SK) classification winner ~\cite{montypython}, and melanoma (M) classification winner ~\cite{recod},  respectively, are shown in Table ~\ref{table:classresults}, as well as 3 fusion strategies ~\cite{jaadarticle}: score averaging (AVGSC), linear SVM (L-SVM), and non-linear SVM (NL-SVM) using a histogram intersection kernel. Fusion strategies utilize all submissions on the final test set, and are carried out via 3-fold cross-validation. SVM input feature vectors included all disease category predictions. Both SVM methods used probabilistic SVM score normalization, producing an output confidence between 0.0 and 1.0 (with 0.5 as binary threshold), correlating with the probability of disease on a balanced dataset ~\cite{codellajrd}.  ROC curves for the 3 submissions and the best fusion strategy (Linear SVM) are shown in Fig. ~\ref{fig:part3roc}.

The 5 major trends observed involve the following: 1) All top submissions implemented various ensembles of deep learning networks. All used additional data sources to train, either from ISIC  ~\cite{casio,recod}, in-house annotations \cite{montypython}, or external sources ~\cite{recod}. 2) Classification of seborrheic keratosis appears to be an easier task in this dataset, compared to melanoma classification. This may reflect aspects of the disease, or bias in the dataset. The best performance came from the team that added additional weakly labelled pattern annotations to their training data ~\cite{montypython}.  3) The top average performer was not the best in any single classification category. 4) The most complex fusion approach (NL-SVM) led to a decrease in performance, whereas simpler methods led to overall improvements in performance, consistent with previous findings ~\cite{jaadarticle}. This challenge is the second benchmark to demonstrate that a collaborative among all participants outperforms any single method alone. 5) Not all thresholds balanced sensitivity and specificity. Probabilistic score normalization in fusions is effective at balancing sensitivity and specificity ~\cite{codellajrd,jaadarticle}.

\section{DISCUSSION \& CONCLUSION}
\label{sec:conclusion}

The International Skin Imaging Collaboration (ISIC) archive was used to host the second public challenge on “Skin Lesion Analysis Toward Melanoma Detection” at the International Symposium on Biomedical Imaging (ISBI) 2017. The challenge was divided into 3 tasks:  segmentation, feature detection (4 classes), and disease classification (3 classes). 2000 images were available for training, 150 for validation, and 600 for testing.  The challenge involved 593 registrations, 81 pre-submissions, and 46 finalized submissions,  making it the largest standardized and comparative study in this field.

Analysis of segmentation results suggest that the average Jaccard Index may not accurately reflect the number of images where automated segmentation falls outside inter-observer variability. Future challenges may adjust the evaluation metric based on this observation. For example, a binary error may be more appropriate (segmentation failure or success), computed by either using multiple segmentations per image to determine a segmentation difference tolerance threshold, or by choosing a fixed threshold as an estimator based on prior studies. 


Poor participation was noted in dermoscopic feature detection; however, submitted systems achieved reasonable performance. Future challenges may adjust the technical format of the task to more closely align with existing image detection benchmarks to better facilitate ease of participation. For example, the output can be formatted as a segmentation or bounding-box detection task.


Analysis of the classification task demonstrates that ensembles of deep learning approaches and additional data led to the highest performance. In addition, collaborative fusions of all participant systems outperformed any single system alone. With the exception of ~\cite{montypython}, submitted methods generate little human interpretable evidence of disease diagnosis. Future work or challenges may give more focus to this need for proper integration into clinical workflows.


Limitations of this study included dataset bias (not all diseases, ages, devices, or ethnicities were represented equally across categories), and incomplete dermoscopic feature annotations. Reliance on single evaluation metrics rather than combinations may also be a limitation ~\cite{metricfusion}. Future challenges will attempt to address these issues in conjunction with the community.


\bibliographystyle{IEEEbib}

\end{document}